\documentclass[11pt]{article}
\usepackage{hyperref,url}
\usepackage{multirow,array}
\usepackage{amsfonts,amsmath,amssymb,amsthm}
\usepackage{rotating,graphicx}
\usepackage{mathptmx,booktabs}
\usepackage{pdflscape}
\usepackage{authblk}
\usepackage[misc]{ifsym}
\usepackage{enumitem}
\setlist{itemsep=0pt}
\usepackage[height=25cm,width=18.75cm]{geometry}

\usepackage{tikz,xcolor,hyperref}

\definecolor{lime}{HTML}{A6CE39}
\DeclareRobustCommand{\orcidicon}{
\hspace{-3mm}	
	\begin{tikzpicture}
	\draw[lime, fill=lime] (0,0) 
	circle [radius=0.16] 
	node[white] {{\fontfamily{qag}\selectfont \tiny ID}};
	\draw[white, fill=white] (0.0,0.0) 
	circle [radius=0.000];
	\end{tikzpicture}
	\hspace{-3mm}
}

\foreach \x in {A, ..., Z}{%
	\expandafter\xdef\csname orcid\x\endcsname{\noexpand\href{https://orcid.org/\csname orcidauthor\x\endcsname}{\noexpand\orcidicon}}
}

\usepackage{subcaption}
\usepackage{adjustbox}
\usepackage{lscape}
\title{Cost-Sensitive Stacking: an Empirical Evaluation}
\date{\empty}

\author[1]{Natalie Lawrance\thanks{Email: natalie.lawrance@vub.be}\orcidA{}(\Letter)}  

\author[1]{Marie-Anne Guerry\orcidB{}} 

\author[2]{George Petrides\orcidC{}} 

\affil[1]{Department of Business Technology and Operations, Vrije Universiteit Brussel (VUB), Brussels, Belgium}

\affil[2]{Department of Mathematics and Statistics, University of Cyprus, Nicosia, Cyprus}

\begin{document}
\maketitle

\vspace{-2mm}
\begin{abstract}
    
Many real-world classification problems are cost-sensitive in nature, such that the misclassification costs
vary between data instances.
Cost-sensitive learning adapts classification algorithms to account for differences
in misclassification costs.
Stacking is an ensemble method that 
uses predictions from several classifiers as the training data for another classifier, which in turn makes the final
classification decision.

While a large body of empirical work exists where stacking is applied in various
domains,  very few of these works take the misclassification costs into account.
In fact, there is no consensus in the literature as to what cost-sensitive stacking is.
In this paper we perform extensive experiments with the aim of establishing what the appropriate
setup for a cost-sensitive stacking ensemble is.
Our experiments, conducted on twelve datasets from a number of application domains, using real,
instance-dependent misclassification costs, show that for best performance,
both levels of stacking require cost-sensitive classification decision.
\end{abstract}

\maketitle

\section{Introduction}

Cost-sensitive learning is relevant in many real-world classification problems,
where different misclassification errors incur different costs.
A prominent example is the field of medicine, where misdiagnosing an ill patient for a healthy one (a false negative) 
entails delayed treatment and potentially life-threatening consequences, while an error in the opposite direction 
(a false positive) would incur unnecessary medical examination costs and stress for the patient. 
Cost-sensitive classifiers can account for the differences in costs not only between different classes, but also
between data instances, making instance-dependent cost-sensitive classification decisions.

Many cost-sensitive classifiers employ ensemble methods, which combine predictions from several classifiers to obtain better
generalisation performance. Superiority of ensembles over individual classifiers
is very well known and has been extensively studied (\cite{Delgado2014DoWN,EnsembleMethodsbook}).
Most cost-sensitive classification ensembles are homogeneous in nature, meaning their components are instantiated
using the same learning algorithm.

\emph{Stacked generalization} or \emph{stacking}~\cite{wolpert1992stacked} is a well known and widely applied heterogeneous ensemble,
where the predictions of classifiers produced by different learning algorithms (the \emph{base-learners})
are used as training inputs to another learning algorithm (the \emph{meta-learner})
to produce a \emph{meta classifier}, which makes the final classification decision.
In the literature, the \emph{base-} and \emph{meta-} levels of stacking are also referred to \emph{level-0} and
\emph{level-1}.

Homogeneous cost-sensitive ensembles such as cost-sensitive boosting and bagging are widely studied 
and have been shown very successful~\cite{petrides2022survey}.
Examples of cost-sensitive stacking, on the other hand, are scarce and unsystematic, representing
for the most part applications to single domains, where the classifiers are trained on synthetic, class-dependent costs 
and are evaluated with cost-insensitive performance metrics.
For a discussion on the importance of real costs for a proper evaluation see the work by~\cite{petrides2022survey}.
In fact, there is currently no consensus as to how a cost-sensitive stacking ensemble is to be composed and at what stage
(level-0 or level-1) cost-sensitive decision-making should be used. 
This can be clearly seen in Table~\ref{tab:litreview}, which gives an overview of existing cost-sensitive stacking
literature.
Stacking is typically made cost-sensitive simply through the application of a cost-sensitive classifier
either at level-0 (CS-CiS), level-1 (CiS-CS) or at both levels of the the ensemble (CS-CS), resulting in a total of three possible stacking setups. To the best of our knowledge, no comparison of all three setups on multiple domains with appropriate evaluation
exists in the literature.
Previous related work used arbitrary artificial costs in model training and evaluated cost-sensitive models
using performance metrics that are either cost-invariant or that focus on the performance of only the positive class.

In this work we aim to fill this gap by providing a thorough comparison of 
various cost-sensitive stacking ensembles on multiple domains using
real, instance-dependent costs and performance metrics appropriate for cost-sensitive problems.

\subsection{Our contributions}
\begin{itemize}
    \item The main contribution of this work is a rigorous empirical comparison of different setups of cost-sensitive stacking ensembles over multiple domains.
     We evaluate using appropriate performance metrics and attempt to establish best practice.
    \item Secondly, we introduce a novel cost-sensitive classifier combination method, inspired by MEC-voting and stacking,
which we call \emph{MEC-weighted-stacking}.
    \item Finally, we present a list of publicly available datasets with clearly defined instance-dependent misclassification costs.
    The costs are based either on the literature, or are defined by us 
    based on both the literature and expert knowledge of the data providers. 
    We also define instance-dependent costs for a well known `credit-g' 
    dataset from the UCI Machine learning repository, for which only class-dependent 
    costs were available to date.
\end{itemize}

\subsection{Outline} 
The remainder of the paper is structured as follows.
Section~\ref{sec:litreview} presents an overview of the relevant literature.
MEC-weighted stacking is introduced in Section~\ref{sec:newMethod}.
Our hypotheses to be tested, the experimental setup and the datasets used in the study are discussed in Section~\ref{sec:Experiments}.
Section~\ref{sec:Results} details the results of our extensive experiments, while the main outcomes and limitations are discussed in Section~\ref{sec:Discussion}.
Section~\ref{sec:Conclusion} concludes the paper.

\section{Related work}\label{sec:litreview}

While stacking has been widely used in machine learning applications (the interested reader is invited to peruse the survey on
stacking literature by~\cite{sesmero2015generating}), 
few works are dedicated to the study of cost-sensitive stacking.

We identified in the literature three different cost-sensitive stacking setups: CiS-CS, CS-CiS or CS-CS, where
the ensemble was made cost-sensitive simply through the application of a cost-sensitive classifier
either at level-0, level-1 or at both levels of the ensemble.
In most cases, the method used to make the classification cost-sensitive is the direct cost-sensitive decision as introduced by~\cite{zadrozny2001learning}, also 
called DMECC~\cite{petrides2022survey}.

One of the first papers to discuss stacking in a cost-sensitive context was \cite{CameronJones2001StackingFM}.
The authors propose cost-insensitive level-0 and cost-sensitive level-1 stacking setup (\emph{CiS-CS setup}), which 
was compared to a number of different classifier combinations schemes on 16 classification problems.
The misclassification costs they used were artificially generated by randomly and uniformally sampling costs from
on the interval $[1,10]$.
Several other studies followed adopting the same CiS-CS stacking setup, however none of the studies 
explicitly  reasoned or justified this choice.

    \begin{table}
    \caption{Summary of cost-sensitive stacking literature}\label{tab:litreview}
 \begin{center}    
    \begin{adjustbox}{scale=0.7}
        \begin{tabular}{lcccccc}
\toprule
\textbf{Publication} &\textbf{    Stacking} & \textbf{Level-0}   & \textbf{Level-1}   & \textbf{ Real} & \textbf{Costs}&  \textbf{CS} \\
                              &       \textbf{setup} &\textbf{algorithm} & \textbf{algorithm} & \textbf{costs} & \textbf{type} & \textbf{ evaluation }\\
\midrule
 \cite{CameronJones2001StackingFM} &               CiS-CS &             DT, KNN, NB &                   LR &            &          c &$\checkmark$ \\
   \cite{Kotsiantis2008StackingCS} &               CS-CiS &             DT, KNN, NB &                   MT &            &          c & \\
         \cite{bahnsen2019fraud} &                CS-CS &                      DT &                   LR &           $\checkmark$& i &$\checkmark$ \\
       \cite{Yan2018ClassifyingID} &               CiS-CS &             DT, KNN, NB &                   LR &            &          c & \\
         \cite{cao2018imcstacking} &               CiS-CS & ExT, GBDT, LDA, LR, RF   &                   LR &            &          c & \\
            \cite{xiong2021cancer} &               CiS-CS &           DT, KNN, RF, SVM &        DT, KNN, NB, SVM &            &          c & \\
          \cite{eivazpour2021cssg} & CiS-CS, CS-CiS, CS-CS &           DT, NB, KNN, SVM &               LR, ExT &            &          c & $\checkmark$ \\
                        \emph{this paper} & CiS-CS, CS-CiS, CS-CS &           DT, KNN, LR, SVM & Adab, DT, KNN, LR, RF, SVM      &$\checkmark$ & i &  $\checkmark$ \\
\midrule
\emph{Costs type:} &\multicolumn{6}{l}{c: class-dependent, i: instance-dependent}\\
\emph{Algorithms:} &\multicolumn{6}{l}{Adab: Adaboost, DT:decision tree, ExT: extremely randomised trees, GBDT: gradient boosted trees, KNN: k-nearest neighbour,} \\
                   &\multicolumn{6}{l}{LDA: linear discriminant, LR: logistic regression, MT: Meta Decision Trees, NB: naive bayes, RF: random forest, SVM: support vector machines} \\
\bottomrule
\end{tabular}

    \end{adjustbox}
  \end{center}
    \end{table}

 Several more papers demonstrated similar
examples of multiple-domain studies of CiS-CS stacking with arbitrary costs (~\cite{cao2018imcstacking, xiong2021cancer, Yan2018ClassifyingID}).
These mainly differ in the type and the number of algorithms that are employed in the ensemble.
We note that all of them used cost-insensitive metrics for classifier evaluation.

~\cite{Kotsiantis2008StackingCS} considers a stacking setup, where level-0 classifiers were cost-sensitive while
level-1 was cost-insensitive (\emph{CS-CiS setup}). 
The misclassification costs were assumed to be equal to the inverse of the class priors.
This approach is very commonly adopted in the absence of information about real misclassification costs.
It is, however, not appropriate, see~\cite{petrides2022survey} for a discussion. 
The resulting stacking classifier was compared to known ensemble methods using classification accuracy,
a metric that by design assumes equal misclassification costs.

Most examples of stacking use different learning algorithms in level-0, however
in his original work Wolpert suggested that this must not be the case and the technique 
can also be applied when a single algorithm is considered.
~\cite{bahnsen2019fraud} propose a cost-sensitive variant of \emph{bag-stacking}, a method originally proposed by~\cite{Ting1997StackingBA},
using bagged cost-sensitive decision trees in level-0 and using cost-sensitive logistic regression in level-1,
thus implicitly proposing a \emph{CS-CS} stacking setup.
To the best of our knowledge, this study is the only example where real instance-dependent costs were used in model training.
Models were evaluated using a cost-sensitive metric called the savings score, proposed in \cite{bahnsen2014example}.

The only study to date that considers all three different cost-sensitive stacking setups is one by~\cite{eivazpour2021cssg}
on the application domain of software defect prediction.
The misclassification costs were selected based on a literature
however the authors emphasised that they treated costs as one of the hyperparameters of the classifier, 
which, we must note, is incorrect, as was previously discussed in~\cite{hand2009measuring}.
The experiments are run on 15 datasets using the same class-dependent cost matrix on all.
Balanced error-based metrics were used for evaluation together with cost-based evaluation metrics.

Identifying real misclassification costs is a complex task, which for many applications may prove too difficult to define and compute.
Most studies resort to artificially generated misclassification costs (see~\cite{petrides2022survey} for a discussion on why this is inappropriate) and error-based evaluation metrics are
typically employed to assess generalisation performance of cost-sensitive stacking.
Examples of metrics used include the AUC, the arithmetic or geometric mean of class-specific accuracies, the F-measure,
and the Matthew's correlation coefficient (MCC).
All of these metrics assume equal misclassification costs, and the F-measure does not incorporate the performance
on the negative class, so using these metrics is not compatible with cost-sensitive learning~\cite{hu2014study}.

One of the challenges of stacking is the choice of the learning algorithms for the ensemble.
Earlier studies proposed to use linear regression to combine level-0 inputs~\cite{ting1999issues}, however
Wolpert does not impose any particular restrictions on which algorithm to use in level-1,
and he believed that his famous `No Free Lunch Theorem'~\cite{wolpert1997no}
applies to the meta-learner as well.
For the overview of which learning algorithms were used in cost-sensitive stacking ensembles to date
we refer our reader to the summary Table~\ref{tab:litreview}.

\section{MEC-weighted stacked generalization}\label{sec:newMethod}

In the typical supervised classification framework, 
a \textit{learning algorithm} $A$ 
is presented with a set $\mathcal{S}$ of data instances $(\textbf{x}_i, y_i)$, each describing some object $i$.
 We call $\textbf{x}_i$ a feature vector, and $y_i$ the class label of that object,
 drawn from a finite, discrete set of classes $\{1,\ldots, K\}$.
 In this paper we will consider the binary classification problem, where $y_i \in \{0,1\}$.
The learning algorithm $A$, given $\mathcal{S}$ as input, after a process called training,
produces a classifier $C$,
whose task is to predict the correict class label
$\hat{y}_C(\textbf{x}_j) \in \{0,1\}$  for a previously unseen feature vector $\textbf{x}_j$.

Training any number $L$ of learning algorithms on the same set of data instances $\mathcal{S}$,
we obtain a set of classifiers $\mathcal{C} = \{C_1 \ldots C_L\}$, and
for each feature vector $\textbf{x}_i$ the corresponding set of predictions $\hat{\mathcal{Y}}(\textbf{x}_i) = \{\hat{y}_{C_1}(\textbf{x}_i), \ldots, \hat{y}_{C_L}(\textbf{x}_i)\}$.
$C$ is called \emph{an ensemble of classifiers} if the predictions from $\hat{\mathcal{Y}}(\textbf{x}_i)$ are combined, in some way, into a single prediction of the class label for the data instance $\textbf{x}_i$.

Stacking differs from other classifier ensembles in that the predictions from
the set $\hat{\mathcal{Y}}(\textbf{x}_i)$ are combined with the original class label $y_i$ to form the set $\mathcal{S}_{meta} =\{(\hat{y}_{C_1}(\textbf{x}_i),\ldots, \hat{y}_{C_L}(\textbf{x}_i)), y_i\}$ of meta level data instances subsequently used in another round of algorithm training to produce a new classifier, which is used to 
obtain the final predictions.

The novel method we propose in this paper is inspired by the cost-sensitive weights for model
 votes paradigm described in~\cite{petrides2022survey}, and consequently called \emph{MEC-weighted stacking}.
To each classifier $C$, we can assign a weight $w_C$ based on that classifier's cost-performance on
the validation set: $w_C = f(\epsilon)$, where $\epsilon$ is the sum of the misclassification costs of all data instances incorrectly classified by $C$ on a validation set
and $f(\epsilon)$ is
a transformation function, which for example can take one of the following forms:
$f(\epsilon) =\text{ln}((1-\epsilon)/\epsilon)$,
$f(\epsilon) =1-\epsilon$,
$f(\epsilon) =\exp((1-\epsilon)/\epsilon)$, and
$f(\epsilon) =((1-\epsilon)/\epsilon)^2$.

The general stacking procedure is thus modified with the additional step of collecting the MEC-weights for each of the predictions 
from the set $\hat{\mathcal{Y}}(\textbf{x}_i)$, yielding the weighted set of predictions
$\hat{\mathcal{Y}}_{MEC}(\textbf{x}_i) = \{(w_{C_1}\hat{y}_{C_1}(\textbf{x}_i),\ldots, w_{C_L}\hat{y}_{C_L}(\textbf{x}_i)), y_i\}$,
which is used in meta classifier training instead of $\hat{\mathcal{Y}}(\textbf{x}_i)$.

\section{Experimental setup}\label{sec:Experiments}

\subsection{Data}
\begin{table}
\caption{Characteristics of the datasets used in our experiments}\label{tab:dataset_description}
\begin{center}
\begin{adjustbox}{width=0.8\textwidth}
	\begin{tabular}{llllrrc}
\toprule
{} &\textbf{Application     } &  \textbf{        Dataset alias} & \textbf{\# instances} & \textbf{\# Attr} & \textbf{ \% positives }&\textbf{Instance-dependent}\\
{} &        \textbf{ domain}  &                                 &                       &                   &                        & \textbf{costs source}\\

\midrule
1  &  Bankruptcy &                           bankruptcy (private) &          404999 &  221    &         3.31 &               this publication \\
2  &      Churn &               churn\_kgl (Kaggle*) &            7043 &    21&        26.54 &        \cite{petrides2022survey} \\
3  &      Churn &                  churn\_AB \cite{bahnsen2015novel} &            9410 &    45&         4.83 &        \cite{bahnsen2015novel} \\
4  &            Credit risk &                   credit\_kgl (Kaggle*) &          112915 &    15&        11.70 &      \cite{bahnsen2014example} \\
5  &            Credit risk &                     credit\_de\_uci \cite{UCI_repo} &            1000 &    20&        30.00 &         this publication  \\
6  &            Credit risk &          credit\_kdd09 \cite{theeramunkong2010new} &           38938 &    39&        19.89 &      \cite{bahnsen2014example} \\
7  &            Credit risk &             credit\_ro\_vub \cite{petrides2020cost} &           18918 &    24&        16.95 &        \cite{petrides2020cost}\\
8  &       Direct marketing &           dm\_pt\_uci \cite{UCI_repo, moro2014data} &           45211 &    17&        11.27 &    \cite{bahnsen2014improving}\\
9 &       Direct marketing &                          dm\_kdd98 \cite{UCI_repo} &      95412 (train) &   479&         5.08 &        \cite{petrides2022survey} \\
   &                     &                                                &     96367 (test) &      &         5.06                          \\
10 &        Fraud detection &              fraud\_ulb\_kgl \cite{le2004machine} &          284807 &    31&         0.17 &        \cite{petrides2022survey}\\
11 &        Fraud detection &        fraud\_ieee\_kgl (Kaggle*) &          590540 &   432&         3.50 &        \cite{petrides2022survey} \\
   &                     &                                                &              &      &           &                                     \\
12 &           HR analytics &                          absenteeism\_be (private) &      36853 (train) &    71&        14.50 &  \cite{lawrance2021predicting}\\
   &                     &                                                &     35884 (test) &      &        10.76 &                                     \\
\midrule
\multicolumn{7}{l}{* Kaggle: https://www.kaggle.com/}\\
   \bottomrule
\end{tabular}
\end{adjustbox}
\end{center}
\end{table}
In this study we use a collection of 10 publicly available datasets and 2 private datasets, for which
misclassification costs have either already been defined or will be defined here.
This collection of datasets represents a number of application domains: \emph{credit scoring},
\emph{customer churn prediction}, \emph{direct marketing}, \emph{credit card fraud detection}, and \emph{HR analytics}.

\subsection{Misclassification costs}
Table~\ref{tab:dataset_description} presents the references both to the datasets and to relevant publications
where the instance-dependent misclassification costs for a given domain were introduced.
Most of the datasets are large, the number of instances ranging between 1000 and almost 600000.
The number of input features ranges from 15 to 479.
All of these datasets demonstrate a large degree of class imbalance, where the percentage of positives reaches at most
30\%, and in dataset \emph{fraud\_ulb\_kgl} less than 1\%.

In this work we propose instance-dependent costs for these two datasets, for which no costs were previously defined.

The German credit dataset is well known and is referred to as  \emph{credit\_de\_uci} in Table~\ref{tab:dataset_description}. 
Only class-dependent costs were available for this data set, where
the prediction task is to identify customers that will default on their loan.
We define instance-dependent costs using the conceptual framework proposed by~\cite{bahnsen2014example}.
For any data instance $i$, the cost of a false negative $C_{FN}^i$ is defined as loss given default and
constitutes 75\% of the credit line,
while the cost of a false positive $C_{FP}^i$ is the loss of the potential profit from rejecting a good customer,
plus the sum of the average expected loss and the average expected profit estimated on the training sample.
We define profits as simply the interests earned on the credit line in the current year. The profits are calculated
using historic interest rates for the year 2000 in Germany, which we apply randomly and uniformly to the whole sample.

The bankruptcy dataset was provided by the credit risk department of a European utilities-provider, who
was interested in predicting the risk of corporate bankruptcy for new customers.
With minor modifications, it readily transfers to the same credit risk model described above.
Here the credit line is equivalent 90 days of utilities usage by the customer, which, in case of default, 
the provider loses in full, so $C_{FP}^i$ equals the credit amount.
The profit margins were provided to us and are calculated per customer based on the assumption of a 12-month contract.
Thus, the $C_{FP}^i$ then equals the annual profit margins for the potentially good customer plus the
expected average loss and expected average profit calculated on the given sample.

\subsubsection{Data preprocessing}
We take care to employ the same preprocessing steps for each of the datasets in the sample, 
as recommended by the works that first published them. 

In addition to that, we apply the following preprocessing steps to all datasets.
All numeric variables are rescaled using the quantile statistics, which are robust to outliers.
Missing values of numeric variables are imputed with sample median, and of 
categorical variables are encoded as a separate category.
All categorical variables are transformed using weight-of-evidence coding~\cite{agterberg1989systematic}.

\subsubsection{Data partitioning} 
The classifier performance estimates are obtained by means of repeated stratified k-fold cross-validation.
The $5 \times 2$ cross-validation suggested by~\cite{Dietterich1998ApproximateST} is used to train and evaluate
stacking ensembles. This resampling is repeated 5 times using different random seeds, and the results are averaged across folds and across iterations.
Large datasets with more than 100000 observations, to keep training times manageable,
were split into five disjoint subsets, uniformly at random.

We note that two datasets in our sample are provided with a separate test set, used to evaluate model performance.
In this case, for fairness of comparison, we perform the split into folds on each of the
training and test datasets using the same seed, we then proceed using the training partition of the
training set and the test partition of the test set. The training partition of the test set remains unused in evaluation.
When training and test data sets contain the same observations at different time periods (e.g.\ in bankruptcy prediction)
we ensure that training and test datasets are disjoint and do not contain overlapping data instances.

\subsection{Learning algorithms}

The choice of the algorithms for the base- and meta-level of stacking remains one of the challenges of
stacked generalization.
To the best of our knowledge, no study exists that demonstrates the
necessity to use a specific algorithm combination in either base- or meta-level of stacking.
The main requirement for the base classifiers of any ensemble is that they are sufficiently accurate (meaning
they predict better than a random guess) and sufficiently diverse (meaning their errors are uncorrelated)~\cite{Dietterich2000EnsembleMI}.
In a heterogeneous ensemble, where the decisions of different learning algorithms are combined,
the number of base-learners need not be large~\cite{rokach2009taxonomy}.
All algorithms below have previously been described and discussed in detail in a number of machine learning textbooks, for example~\cite{hastie2009elements},
so we refrain from repeating these descriptions here.

\subsubsection{Base-learners}
The base learners in our experiments are four well known classification algorithms, which are: CART Decision Tree (DT), K-Nearest Neighbors (KNN),
Support Vector Machines (SVM) and Logistic Regression (LR).
Unlike~\cite{cao2018imcstacking} and~\cite{xiong2021cancer} before us, we choose not to use ensembles such as Random Forest or
Extremely Randomised Trees in the base level of stacking. The reasons for this are two-fold.
Firstly, ensembles in general, and stacking in particular are typically built on weak base-learners, which these very powerful models, which are themselves ensembles, certainly are not.
Secondly these methods are based on decision trees
and their errors will be correlated with DT.\@
In our choice we also considered the recommendations of~\cite{Delgado2014DoWN},
one of the largest empirical studies known to date comparing algorithm performance on 
121 datasets. Their results on binary problems (55 UCI datasets) demonstrate that Random Forest, SVM, Bagging and Decision
Trees have the highest probability of obtaining more than 95\% of accuracy, while classifiers of the Naive Bayes (NB) family are not competitive in comparison.
We therefore do not include NB in our experiments, unlike some previous studies in cost-sensitive stacking. 

\subsubsection{Meta-learners}
The choice of the meta-learner constitutes a challenge as well, 
as was called ’the black art’ by the original author of stacked generalization~\cite{wolpert1992stacked}.
To keep the scale of our experiments manageable and to allow for statistical comparison between stacking and base classifiers,
we use the same four algorithms that were used in the level-0 of stacking. In addition to that we also use two
homogeneous ensemble methods that, according to~\cite{Delgado2014DoWN}, perform well on most problems, namely
Adaboost (Ada) and Random Forest (RF).

\subsubsection{Cost-sensitive learners}

While many variants of cost-sensitive learning algorithms exist that can incorporate the misclassification costs during classifier training~\cite{petrides2022survey},
in this study we are not interested in comparing cost-sensitive learning algorithms, but in
ways of combining cost-sensitive and cost-insensitive learners in a single ensemble.
For our purposes it is important that the two classifiers we compare are different in all but one thing,
that is the composition of the ensemble.
We therefore choose to turn known cost-insensitive classifiers cost sensitive by applying a cost-sensitive threshold adjustment method called
DMECC~\cite{petrides2022survey}. In this method, each data instance is classified according to its individual cost-sensitive decision threshold,
which is based on the ratio of misclassification costs of that particular data instance.
The threshold is calculated as follows: $ T_{cs}^i = \frac{C_{FP}^i - C_{TN}^i}{{C_{FP}^i - C_{TN}^i + C_{FN}^i - C_{TP}^i}}\enspace$,
where $C_{TN}^i$ and $C_{TP}^i$ refer to the costs of correct classification,
and $C_{FN}^i$ and $C_{FP}^i$ refer to the misclassification costs of the positive and negative data instances respectively.
A given record is classified as positive when its estimated probability of being positive exceeds its individual cost-sensitive
threshold $T_{cs}^i$~\cite{zadrozny2001learning}.

Since some learning algorithms (such as DT or SVM) are known not to produce reliable probability estimates,
we applied isotonic calibration~\cite{zadrozny2002transforming} to all base-learners.

\subsubsection{Cost-sensitive stacking}
To the best of our knowledge no definition exists of what constitutes
cost-sensitive stacking.
Based on the insights from the literature earlier discussed in Section~\ref{sec:litreview},
we see three main possibilities of introducing cost-sensitivity into 
the ensemble structure.

\begin{enumerate}
\item 
Level-0 classifiers are cost-sensitive,
level-1 classifiers are cost-insensitive.
\item 
Level-0 classifiers are cost-insensitive,
level-1 classifiers are cost-sensitive.
\item Both level-0 and level-1 classifiers are cost-sensitive.
\end{enumerate}

We consider 4 functional forms for the MEC-weights as introduced in Section~\ref{sec:newMethod},
which resulted in a total of 15 stacking setups to be compared.
The complete list of ensemble compositions is presented in Table~\ref{tab:stacking_setups}.

\begin{table}[!htp]
\caption{The complete list of stacking setups compared in our study.}\label{tab:stacking_setups}
    \begin{adjustbox}{scale=0.9, center}
        \begin{tabular}{lllll}
    \toprule
    &\textbf{Stacking setup} & \textbf{Level-0} & \textbf{Level-1 } & \textbf{Level-1} \\
    &\textbf{alias} & \textbf{algorithm type} & \textbf{input weights $f(\epsilon)$} & \textbf{algorithm type} \\
    \midrule
    1&type-1                         & CS                             & 1                                                       & CiS                            \\
    2&type-1\_exp                    & CS                             & $\exp((1-\epsilon)/\epsilon)$                           & CiS                            \\
    3&type-1\_ln                    & CS                             & $\ln((1-\epsilon)/\epsilon)$                      & CiS                            \\
    4&type-1\_sq                     & CS                             & $((1-\epsilon)/\epsilon)^2$                             & CiS                            \\
    5&type-1\_acc                    & CS                             & $1-\epsilon$                                            & CiS                            \\
    6&type-2                         & CiS                            & 1                                                       & CS                             \\
    7&type-2\_exp                    & CiS                            & $\exp((1-\epsilon)/\epsilon)$                           & CS                             \\
    8&type-2\_ln                    & CiS                            & $\ln((1-\epsilon)/\epsilon)$                      & CS                             \\
    9&type-2\_sq                     & CiS                            & $((1-\epsilon)/\epsilon)^2$                             & CS                             \\
    10&type-2\_acc                    & CiS                            & $1-\epsilon$                                            & CS                             \\
    11&type-3                         & CS                             & 1                                                       & CS                             \\
    12&type-3\_exp                    & CS                             & $\exp((1-\epsilon)/\epsilon)$                           & CS                             \\
    13&type-3\_ln                    & CS                             & $\ln((1-\epsilon)/\epsilon)$                      & CS                             \\
    14&type-3\_sq                     & CS                             & $((1-\epsilon)/\epsilon)^2$                             & CS                             \\
    15&type-3\_acc                    & CS                             & $1-\epsilon$                                            & CS                            \\
    \bottomrule
\end{tabular}
    \end{adjustbox}
\end{table}

\textbf{MEC-weighted stacking} renders the level-1 classifier cost sensitive through manipulation of the
training data in a cost-sensitive way by applying MEC-weights to the training data of the level-1 classifier.
We consider it an alternative  to obtaining an ensemble where both training levels are cost-sensitive,
which is the third stacking setup stated above.
\subsubsection{Software used}
All of our experiments were performed using the Python programming language (version 3.8).
Cost-insensitive algorithm implementations came from the \emph{scikit-learn} (version 1.1.1)
Python library~\cite{scikit-learn}, while the cost-sensitive implementations are our own.

\subsection{Evaluation}\label{sec:eval_metrics}

\subsubsection{Evaluation metrics}
Contrary to previous studies in cost-sensitive stacking, we would like to emphasise the importance of 
using appropriate evaluation metrics for cost-sensitive classifiers.
Most authors use traditional evaluation metrics such as ROC\_AUC, Precision or F1 metric.
ROC\_AUC is known to be cost-invariant, since it is a measure that aggregates classifier performance
over all possible class-dependent thresholds, thus implicitly averaging performance over
multiple class-dependent costs, which is not appropriate.
Other error-based metrics typically assume equal class-dependent costs, which, again, is not appropriate,
when instance-dependent costs are known at estimation time.
Cost-sensitive learning aims to adapt the classification decision of a learning algorithm to the
differences between misclassification costs assigned to each of the classes.
It is therefore important that the evaluation metrics used to assess the performance of cost-sensitive
classifiers is also adapted to account for the difference in misclassification costs.
The typical evaluation metric used in cost-sensitive literature is
the total misclassification cost~\cite{elkan2001foundations},
that simply adds up the errors weighted with their individual misclassification costs, as defined on the test set.
Another option is to normalise the total misclassification cost over some budget constraint, which will depend
on the application domain.
A more general way to do this is to use the savings score proposed in~\cite{bahnsen2014example}, where
the total misclassification costs are normalised with the cost of either misclassifying all positives as negatives, 
or misclassifying all negatives as positives, which ever is smallest.
This gives a metric on the interval between 0 and 1, facilitating comparison across different datasets, when necessary.

Since the majority of comparisons in our study is performed based on average ranks,
it requires no commensurability of the evaluation metrics, so the models are ranked according to their total misclassification costs, which allows for more precise outcome.
\subsubsection{Multiple classifier comparison}

In order to compare multiple classifiers on multiple datasets, we use the standard approach of the combination of
the Friedman omnibus test and post-hoc Nemenyi test~\cite{japkowicz2011evaluating}.
The Friedman test is conducted under the null-hypothesis that all algorithms in comparison are equivalent in performance.
If this null-hypothesis is rejected, the post-hoc test can be performed to identify
pairs of classifiers whose performance is significantly different, which is measured using the critical difference
statistic, and can be visualised using the critical differences diagrams~\cite{demvsar2006statistical}.
The non-parametric tests, such as the Friedman test, are preferred in case where the number of datasets
in comparison is less than 30, which is the number of datasets necessary to satisfy the normality assumptions of
parametric statistical tests, such as ANOVA~\cite{demvsar2006statistical}.
The post-hoc test is known to be of low power, not rejecting the null even if the null was rejected for the Friedman test.
In this case, we additionally apply Wilcoxon signed-ranks test, as appropriate, which is used for pairwise comparisons
of classifiers on multiple datasets. This test ranks differences in performances of a given pair of classifiers, 
under the null hypothesis that the median difference in ranks is zero. It therefore allows establishing whether the
observed differences in performance between two classifiers are significant. 
It is considered more powerful than its parametric equivalent, the paired t-test when the assumptions of the latter
cannot be guaranteed. It is also considered more powerful than the
Sign test, which counts the number of wins, losses and ties~\cite{demvsar2006statistical}.

\section{Experimental results}\label{sec:Results}

The purpose of our experiments is twofold.
Firstly, we would like to compare the performance of the
different cost-sensitive stacking setups in order to
determine which of them results in the lowest cost-loss and can be recommended to practitioners.
Secondly, we aim to empirically evaluate MEC-weighted stacking,
which is a new cost-sensitive stacking method we earlier described in Section~\ref{sec:newMethod}.

Despite our best efforts, not all classifiers trained successfully on all 12 datasets.
In particular, we were unable to collect results for the MEC-weighted stacking
where the weights were defined by the logarithmic function on the credit scoring problem \emph{credit\_ro\_vub}, 
and MEC-weighted stacking with exponential weights were missing on the fraud detection dataset \emph{fraud\_ulb\_kgl}.
The full results for all 15 stacking setups are thus available on 10 datasets, instead of 12.
Unweighted stacking results, however, are available on all 12 datasets, which we briefly discuss, for completeness.

\subsection{Finding the best cost-senstive stacking setup}
\subsubsection{Overall comparison} 

\begin{figure}
   \centering
   \includegraphics[scale=0.5]{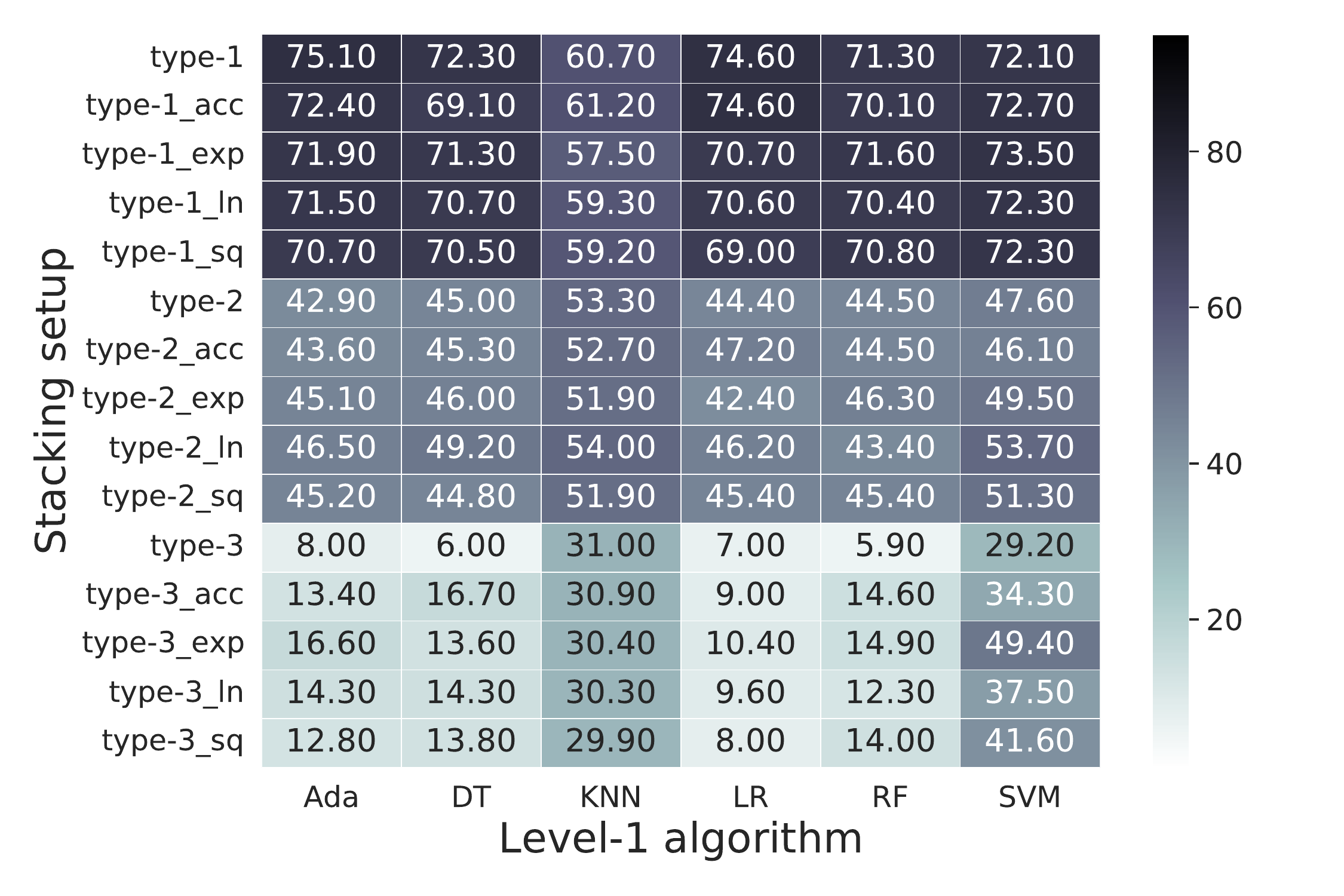}
   \caption[Average ranks]{Comparing all classifiers by average rank across 10 datasets. 
   Lower numbers correspond to better rank.}\label{fig:all_avg_ranks}
\end{figure}

We begin with an overall comparison, where all classifiers are evaluated and ranked on each of the 10 datasets,
and for each of them an average rank is computed across all datasets.
Figure~\ref{fig:all_avg_ranks} presents the average ranks for all stacking classifiers, where the vertical axis
shows the stacking setup and the horizontal axis shows the corresponding level-1 algorithm.
The comparison consists of a total of 90 classifiers (6 algorithms and 15 stacking setups).
For brevity, we adopt the aliases for each of the stacking setups 
earlier presented in Table~\ref{tab:stacking_setups}.

We notice immediately that the ranking demonstrates clusters with 
stacking ensembles of type-3 ranking the best,
while type-1 ensembles rank the worst. 
We note that models built with KNN and SVM algorithms tend to rank lower than decision tree based models or logistic regression.
However, the general picture of type-3 stacking ranking the best and type-1 ranking the worst
remains unchanged for KNN and SVM,
although the differences in ranks between the three groups are smaller than for other algorithms.

Whether these differences in ranks are statistically significant
will be discussed in the following subsection, where we 
demonstrate the outcomes of statistical tests that compare the performance
of various stacking classifiers across multiple domains.

\subsubsection{Comparing unweighted stacking setups on 12 datasets}

\begin{table}
    \caption{The outcome of the Friedman multiple hypothesis testing.}\label{tab:Friedman_test}
    \centering
    \begin{adjustbox}{width=\textwidth, center}
    \begin{tabular}{llllllll}
\toprule
                          & Test statistic ($\chi_{(k-1)}$)   & Ada      & DT       & KNN      & LR       & RF       & SVM      \\
\midrule
Unweighted ($k=3, n=12$)  & 6.5  & 32.69**  & 32.59**  & 32.79**  & 32.69**  & 32.59**  & 32.59**  \\
All ($k=15, n=10$)        & 3.94 & 132.44** & 132.09** & 132.35** & 132.29** & 132.09** & 132.09** \\
\midrule
&\multicolumn{7}{l}{** significant at 0.01 level} \\
&\multicolumn{7}{l}{$k$: number of stacking setups in comparison, $n$: number of datasets in comparison}\\
\bottomrule
\end{tabular}
    \end{adjustbox}
\end{table}

\begin{figure}
    \centering
    \begin{subfigure}[b]{0.31\textwidth}
        \centering
        \includegraphics[width=\textwidth]{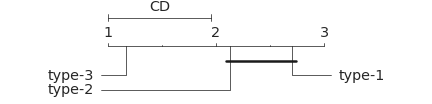}
        \caption{Adaboost}\label{fig:Adaboost}
    \end{subfigure}
    \hfill
    \begin{subfigure}[b]{0.31\textwidth}
        \centering
        \includegraphics[width=\textwidth]{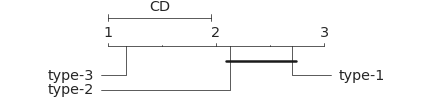}
        \caption{Decision Tree}\label{fig:CD_DT}
    \end{subfigure}
    \hfill
    \begin{subfigure}[b]{0.31\textwidth}
        \centering
        \includegraphics[width=\textwidth]{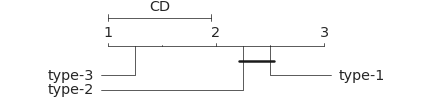}
        \caption{K-Nearest Neighbors}\label{fig:CD_KNN}
    \end{subfigure}

   \bigskip
   
    \begin{subfigure}[b]{0.31\textwidth}
        \centering
        \includegraphics[width=\textwidth]{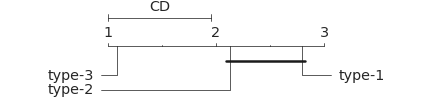}
        \caption{Logistic Regression}\label{fig:CD_LR}
    \end{subfigure}
    \hfill
    \begin{subfigure}[b]{0.31\textwidth}
        \centering
        \includegraphics[width=\textwidth]{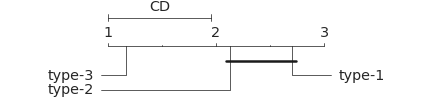}
        \caption{Random Forest}\label{fig:CD_RF}
    \end{subfigure}
    \hfill
    \begin{subfigure}[b]{0.31\textwidth}
        \centering
        \includegraphics[width=\textwidth]{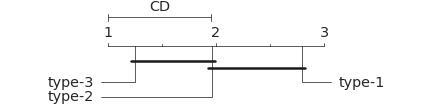}
       \caption{Support-Vector Machine}\label{fig:CD_SVM}
    \end{subfigure}
       \caption{Pairwise comparison of the three unweighted stacking setups on 12 datasets using Nemenyi test at 0.05 significance level.}\label{fig:CD_all_unweighted_setups}
\end{figure}

We begin by testing the null hypothesis that the three unweighted stacking setups show no difference in performance.
The comparison is performed for each of the six classification algorithms used as level-1 learners.
The null hypothesis of the Friedman test was rejected for all 6 comparisons, and the test statistics are presented
in row 1 of Table~\ref{tab:Friedman_test}.

We proceed with the post-hoc Nemenyi test to evaluate the 
alternative hypothesis that the performance of three stackings setups is not equal. 
Figure~\ref{fig:CD_all_unweighted_setups} presents the results of the post-hoc tests at the 0.05 significance level.
We find that type-3 stacking ranks best and is significantly different from both type-2 and type-1 for all algorithms except SVM,
where the difference is only significant for the comparison between type-3 and type-1, but no conclusions can be made
regarding the differences between ensembles of type-3 and type-2.
Similarly, no conclusions can be made regarding the differences in rank between type-2 and type-1 stacking ensembles.

Since the outcome of the post-hoc tests are ambiguous in the case of SVM,
we also perform the Wilcoxon rank sum test under the null hypothesis that the median of the paired differences is zero.
For the comparison between type-3 and type-2 unweighted stacking the null is rejected at 0.05 level.

We conclude from these tests that type-3 stacking performs significantly better than the other two stacking setups.

\subsubsection{Comparing all cost-sensitive stacking setups on 10 datasets}

\begin{figure}
    \centering
    \begin{subfigure}[b]{0.31\textwidth}
        \centering
        \includegraphics[width=\textwidth]{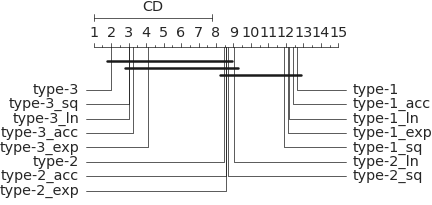}
        \caption{Adaboost}\label{fig:CD_Adaboost1}
    \end{subfigure}
    \hfill
    \begin{subfigure}[b]{0.31\textwidth}
        \centering
        \includegraphics[width=\textwidth]{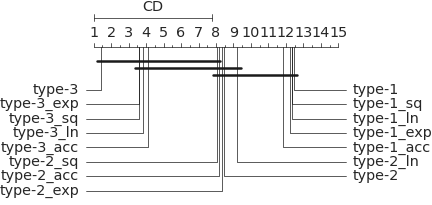}
        \caption{Decision Tree}\label{fig:CD_DT1}
    \end{subfigure}
    \hfill
    \begin{subfigure}[b]{0.31\textwidth}
        \centering
        \includegraphics[width=\textwidth]{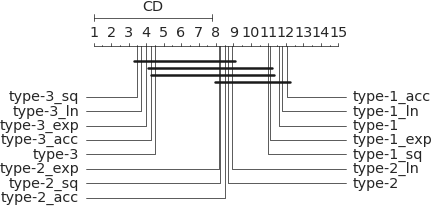}
        \caption{K-Nearest Neighbors}\label{fig:CD_KNN1}
    \end{subfigure}

   \bigskip
   
    \begin{subfigure}[b]{0.31\textwidth}
        \centering
        \includegraphics[width=\textwidth]{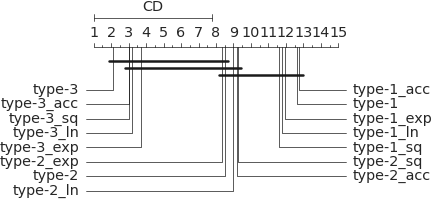}
        \caption{Logistic Regression}\label{fig:CD_LR1}
    \end{subfigure}
    \hfill
    \begin{subfigure}[b]{0.31\textwidth}
        \centering
        \includegraphics[width=\textwidth]{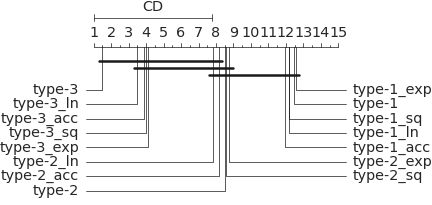}
        \caption{Random Forest}\label{fig:CD_RF1}
    \end{subfigure}
    \hfill
    \begin{subfigure}[b]{0.31\textwidth}
        \centering
        \includegraphics[width=\textwidth]{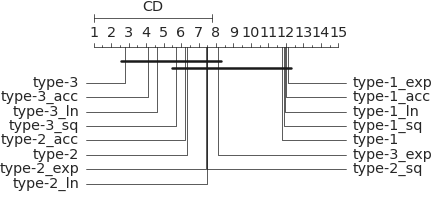}
        \caption{Support-Vector Machine}\label{fig:CD_SVM1}
    \end{subfigure}
       \caption{Comparing all stacking setups on 10 datasets using Nemenyi post-hoc test at 0.05 significance level.
       }\label{fig:CD_all_setups}
\end{figure}
We proceed to compare all 15 stacking classifiers on 10 datasets.
The outcome of the Friedman rank sum test can be found in row 2 of Table~\ref{tab:Friedman_test}.
The null hypothesis of the Friedman test is rejected for every meta-learner at the 1\%
significance level, so we conclude that the performance of all 15 models
is not equal and proceed with the post-hoc test.
Figure~\ref{fig:CD_all_setups} shows the outcome of the Nemenyi test at 0.05 significance level.

These are for the most part consistent with what we observed in Figure~\ref{fig:all_avg_ranks}, where the
classifiers tend to cluster by stacking setup, type-3 being the leader, type-2 the second-best and type-1 ranking worst.
Similar to what we observed above with unweighted stacking, we can reject the null
that type-3 stacking and its MEC-weighted variants are equal in performance to type-1 stacking and variants.
This holds for all algorithms except KNN and SVM. 
For stacking ensembles with KNN in level-1 \emph{type-3} and \emph{type-3\_acc} classifiers
are not significantly different from \emph{type-1\_exp} and \emph{type-1\_sq}, while for SVM no
significant differences were detected between \emph{type-3\_exp} and \emph{type-3\_sq} and other type-1 ensembles.

Since Nemenyi post hoc test is not powerful enough to establish whether the differences between
the three stacking setups are statistically significant, additional testing is required.
From the outcomes of the post hoc test we observed that type-3 stacking generally tends to rank highest,
and is therefore of most interest to us.
We therefore perform the Wilcoxon rank sum test for all combinations of pairwise comparisons of stacking
 algorithms of type-3 vs type-1 and of type-3 vs type-2 under the null hypothesis that the median of the rank differences
 between the two groups is equal to zero.
The complete tables with the obtained test statistics and p-values can be found in the Appendix.
We find that the null could be confidently rejected for all comparisons between type-3 and type-1 stacking ensembles,
we refer the reader to the Table~\ref{tab:wilcoxon_3vs_1} in the Appendix for details.

As for the comparison of stacking type-3 with type-2, the only algorithm where the null 
could not be rejected was SVM. We found that the differences between all type-3 MEC-weighted stacking variants
and type-2 stacking ensembles were not significant.
However, type-3 unweighted stacking was significantly different from all type-2 stacking variants, see Table~\ref{apx:Wilcoxon_type3_vs_type2} for details.

We can therefore recommend type-3 stacking, where both levels of stacking are cost-sensitive, as the winner.

\subsection{Evaluating MEC-weighted stacking}

\begin{table}
    \caption[Wilcoxon test outcomes, comparing unweighted and MEC-weighted stacking]{
        Pairwise comparison of unweighted and MEC-weighted stacking using Wilcoxon rank sum test.
        Statistically significant values are marked with boldface (significance 0.05) and italics (significance 0.1). }\label{tab:Wilcoxon_MEC_vs_stacking}
    \centering
    \begin{adjustbox}{width=0.44\textwidth, center}
            \begin{tabular}{lllll}
        \toprule
    Level-1   &      \multicolumn{4}{c}{Unweighted stacking type-1 vs }                              \\
    algorithm & type-1\_acc & type-1\_exp & type-1\_ln & type-1\_sq \\
   \midrule
    Ada       & 18.0 (0.3)              & 18.0 (0.3)              & 18.0 (0.3)             & 18.0 (0.3)             \\
    DT        & 23.0 (0.63)             & 23.0 (0.63)             & 23.0 (0.63)            & 23.0 (0.63)            \\
    KNN       & 27.5 (1.0)              & 21.0 (0.56)             & 17.0 (0.32)            & 26.5 (0.92)            \\
    LR        & 27.0 (0.96)             & 14.0 (0.15)             & 22.0 (0.56)            & \textit{10.5 (0.08)}   \\
    RF        & 23.0 (0.63)             & 23.0 (0.63)             & 23.0 (0.63)            & 23.0 (0.63)            \\
    SVM       & 18.0 (0.3)              & 27.0 (0.96)             & 22.0 (0.56)            & 27.0 (0.96)            \\
   \midrule
              &      \multicolumn{4}{c}{Unweighted stacking type-2 vs }                                             \\
              &  type-2\_acc            &  type-2\_exp            & type-2\_ln             &  type-2\_sq            \\
   \midrule
    Ada       & 22.0 (0.56)             & 23.0 (0.63)             & 23.0 (0.63)            & 23.0 (0.63)            \\
    DT        & 27.5 (1.0)              & 27.5 (1.0)              & 26.5 (0.92)            & 18.5 (0.35)            \\
    KNN       & 25.0 (0.8)              & 21.0 (0.5)              & 25.0 (0.8)             & 21.0 (0.5)             \\
    LR        & \textit{10.5 (0.08)}    & 20.5 (0.47)             & 24.5 (0.76)            & 16.5 (0.26)            \\
    RF        & 27.5 (1.0)              & 22.5 (0.61)             & 17.5 (0.3)             & 26.5 (0.92)            \\
    SVM       & 23.5 (0.68)             & 19.5 (0.41)             & 18.5 (0.36)            & \textit{10.5 (0.08)}   \\
   \midrule
              &      \multicolumn{4}{c}{Unweighted stacking type-3 vs }                                             \\
              &  type-3\_acc            &     type-3\_exp         &  type-3\_ln            & type-3\_sq             \\
   \midrule
    Ada       & 13.0 (0.16)             & 19.0 (0.43)             & 19.0 (0.43)            & 19.0 (0.43)            \\
    DT        & \textbf{2.0 (0.01)}     & 11.0 (0.11)             & \textbf{0.0 (0.0)}     & \textit{10.0 (0.08)}   \\
    KNN       & 24.0 (0.77)             & 17.0 (0.32)             & 24.0 (0.77)            & 16.0 (0.28)            \\
    LR        & 13.0 (0.16)             & 19.0 (0.43)             & 14.0 (0.19)            & 24.0 (0.77)            \\
    RF        & \textbf{4.0 (0.01)}     & \textbf{8.0 (0.05)}     & \textbf{3.0 (0.01)}    & \textit{9.0 (0.06)}    \\
    SVM       & \textit{10.0 (0.08)}    & \textbf{0.0 (0.0)}      & \textit{9.0 (0.06)}    & \textbf{7.0 (0.04)}  \\ 
   \midrule
\bottomrule   
\end{tabular}
    \end{adjustbox}
\end{table}   

Our next research question is whether within the same setup, MEC-weighted stacking offers any improvement
over the unweighted stacking. 
To determine whether there is a statistically significant difference
in performance between the MEC-weighted stacking models and their unweighted counterparts
we perform pairwise comparison using Wilcoxon rank sum test.
The test statistics and corresponding p-values from the 72 comparisons are reported in Table~\ref{tab:Wilcoxon_MEC_vs_stacking}.
Values that are significant at 5\% level were highlighted with boldface text, weakly significant values at 10\% level
were highlighted with italics.
As previously, the results are reported per learning algorithm used as the level-1 classifier.

We find almost no significant differences in performance between unweighted and weighted stacking for
setups of type-1 and type-2 with rare exceptions.
Surprisingly, only one comparison of stacking type-1, where the level-1 classifier is cost-insensitive shows a significant difference in performance.
Namely, the stacking setup \emph{type-1\_sq} learned with Logistic Regression in level-1.
Referring back to the average rankings reported in Figure~\ref{fig:all_avg_ranks}
it happens to be the best performing Logistic Regression stacking of type-1, so in this instance MEC-weighted
stacking is significantly better than its counterpart with equally weighted meta-inputs.
It is, however, an exception, and we must conclude that introducing
cost-sensitivity through MEC-weights into level-1 of stacking has no positive impact on type-1 stacking performance.

Similar conclusions can be drawn for stacking of type-2, where base classifiers are cost-insensitive but
the meta-classifier is made cost-sensitive using DMECC. 
Here we observe only two statistically significant test outcomes, both of which have lower average ranks
than the unweighted stacking of type-2.

For the setup of type-3 where both level-0 and level-1 of stacking are made cost-sensitive using DMECC,
the null could not be rejected for AdaBoost, Logistic Regression and KNN.
Most of the MEC-weighted ensembles built with Decision Tree, Random Forest and SVM were significantly different from unweighted stacking
of type-3.
Looking at the differences in the average ranks, however we note that unweighted stacking of type-3 ranks noticeably
better than any of the MEC-weighted models.

We must therefore conclude that MEC-weighted stacking does not offer a statistically significant improvement over
conventional stacking.

\subsection{Comparing cost-sensitive stacking with single cost-sensitive models}
Finally, one might ask whether the effort involved in training the level-1 classifier is worth it.
To answer this we compare the best stacking classifier with the corresponding single classifier.
Having previously determined the best stacking setup, where DMECC was applied in both levels, and no MEC-weights are applied,
we will omit other classifiers from this analysis.
We average classifier performance across cross-validation folds using the savings metric (see Section~\ref{sec:eval_metrics})
for commensurability.
We also rank the resulting selection of classifiers by savings and average ranks across 12 datasets.
The results are reported in Table~\ref{tab:stacking_vs_base}, where the winning classifier (per algorithm) is marked
with boldface font, the best performer per dataset is marked with italics.

We note that cost-sensitive stacking always achieves positive savings, meaning its total misclassification costs
are lower than the predetermined budget.
Stacking has higher average savings on all algorithms except KNN and Random Forest.
In terms of average ranks, stacking wins for all level-1 algorithms except KNN, which, we note, is one of the 
worst ranking algorithms in our study.

\begin{table}
    \caption{Comparing single classifiers with type-3 unweighted stacking.
    Savings score is reported for each classifier, higher is better.
    Best model per dataset is marked with italics.
    The winning classifier is marked with boldface letters.}\label{tab:stacking_vs_base}
    \centering
    \begin{adjustbox}{width=\textwidth, center}
        \begin{tabular}{l|cc|cc|cc|cc|cc|cc}
\toprule
                   & \multicolumn{2}{c}{Ada}   & \multicolumn{2}{c}{DT}  & \multicolumn{2}{c}{KNN}         & \multicolumn{2}{c}{LR}          & \multicolumn{2}{c}{RF}          & \multicolumn{2}{c}{SVM} \\
Dataset            & Single         & Stacking       & Single & Stacking       & Single         & Stacking       & Single         & Stacking       & Single         & Stacking       & Single         & Stacking                           \\
\midrule
absenteeism\_be\_1 & 0.188          & \textbf{0.225} & 0.219  & \textbf{0.242} & 0.199          & \textbf{0.224} & 0.188          & \textbf{0.23}  & 0.172          & \emph{\textbf{0.243}} & 0.188          & \textbf{0.223}                     \\
bankruptcy         & 0.03           & \textbf{0.123} & 0.112  & \textbf{0.123} & \textbf{0.105} & 0.058          & -0.043         & \emph{\textbf{0.126}} & \textbf{0.31}  & 0.123          & -0.024         & \textbf{0.024}                     \\
churn\_AB          & \emph{\textbf{0.171}} & 0.081          & 0.052  & 0.087          & \textbf{0.07}  & 0.019          & 0.062          & \textbf{0.082} & \textbf{0.1}   & 0.086          & 0.033          & \textbf{0.04}                      \\
churn\_kgl         & \emph{\textbf{0.311}} & 0.295          & 0.0    & \textbf{0.297} & \textbf{0.242} & 0.031          & \textbf{0.302} & 0.295          & 0.273          & \textbf{0.296} & \textbf{0.225} & 0.066                              \\
credit\_de\_uci    & \textbf{0.399} & 0.391          & 0.293  & \textbf{0.387} & 0.337          & \textbf{0.351} & \textbf{0.396} & 0.388          & \emph{\textbf{0.424}} & 0.386          & \textbf{0.405} & 0.354                              \\
credit\_kdd09      & \emph{\textbf{0.318}} & 0.303          & 0.277  & \textbf{0.302} & \textbf{0.287} & 0.276          & \textbf{0.312} & 0.303          & \textbf{0.313} & 0.302          & \textbf{0.289} & 0.279                              \\
credit\_kgl        & \emph{\textbf{0.511}} & 0.411          & 0.156  & \textbf{0.411} & \textbf{0.408} & 0.202          & -0.053         & \textbf{0.411} & \textbf{0.499} & 0.411          & \textbf{0.378} & 0.175                              \\
credit\_ro\_vub    & 1.793          & \textbf{1.796} & 1.787  & \textbf{1.795} & \textbf{1.786} & 1.785          & 1.727          & \textbf{1.793} & 1.762          & \emph{\textbf{1.797}} & 1.773          & \textbf{1.786}                     \\
dm\_kdd98\_train   & 0.108          & \textbf{0.143} & 0.033  & \emph{\textbf{0.147}} & 0.035          & \textbf{0.061} & \textbf{0.122} & 0.119          & 0.059          & \emph{\textbf{0.147}} & 0.036          & \textbf{0.044}                     \\
dm\_pt\_uci        & \emph{\textbf{0.568}} & 0.558          & 0.537  & \textbf{0.557} & \textbf{0.551} & 0.511          & \textbf{0.562} & 0.558          & 0.556          & \textbf{0.557} & \textbf{0.551} & 0.529                              \\
fraud\_ieee\_kgl   & 0.109          & \textbf{0.494} & 0.444  & \textbf{0.505} & \textbf{0.477} & 0.438          & 0.372          & \textbf{0.495} & \emph{\textbf{0.584}} & 0.506          & 0.407          & \textbf{0.444}                     \\
fraud\_ulb\_kgl    & -0.062         & \textbf{0.714} & 0.625  & \textbf{0.701} & 0.679          & \textbf{0.706} & \textbf{0.75}  & 0.72           &\emph{\textbf{0.762}} & 0.728          & -0.124         & \textbf{0.688}                     \\
\midrule
\textbf{Avg Savings}       & 0.37   & \textbf{0.461} & 0.378  & \textbf{0.463} & \textbf{0.431} & 0.389          & 0.391          & \textbf{0.46}  & \textbf{0.484} & 0.465          & 0.345          & \textbf{0.388}\\
\textbf{Avg Rank}  & 5.08           & \textbf{4.17}  & 9.42  & \textbf{4.33}   & \textbf{8.17}  & 9.17           & 6.75           & \textbf{4.08}  & 4.58           & \textbf{3.83}  & 9.25          & \textbf{9.08}\\
\bottomrule
\end{tabular}
    \end{adjustbox}
\end{table}   

\section{Discussion}\label{sec:Discussion}

\paragraph*{Outcome 1: using cost-sensitive models in both levels of stacking is recommended}

The results presented in this paper, have demonstrated that there is a statistically significant difference
in performance between the three different stacking setups considered in our experiments,
namely \emph{CiS-CS}, \emph{CS-CiS}, and \emph{CS-CS}.
Contrary to the majority of cost-sensitive stacking papers that assumed that one level
of cost-sensitive decision-making is sufficient, 
our experiments demonstrate that stacking models where the DMECC was applied in both levels of stacking
achieved the highest ranking.

While these conclusions hold for this particular post-training method, cost-sensitivity can be introduced
either before or during training of the learning algorithm.
Further experiments are required to investigate how different cost-sensitive methods affect our conclusions.
Now that we have established how cost-sensitive stacking should be built, future work can focus on combining
various kinds of cost-sensitive algorithms, including pre-, during- and post-training cost-sensitive methods~\cite{petrides2022survey}.
Another interesting avenue for future research would be investigating homogeneous cost-sensitive stacking, an example
of which was proposed in~\cite{bahnsen2019fraud} using cost-sensitive decision trees as base classifiers and cost-sensitive
logistic regression as level-1 classifier. 

\paragraph*{Outcome 2: cost-insensitive classifiers do not perform well when costs are known, even in stacking}
As was previously shown in~\cite{lawrance2021predicting}
cost-insensitive classifiers, having no way to account for differences in misclassification costs, 
typically perform worse than cost-sensitive models when evaluated using cost-based performance metrics.
In our study, we observed yet another confirmation
to this in the context of heterogeneous ensembles, where base-learners were cost-sensitive and 
meta-learners were cost-insensitive.
It is, however, surprising that applying MEC-weights has no positive impact on the performance of
these cost-insensitive meta-learners. 
So we conclude that the transfer of cost information via cost-sensitive decision-making of the base-classifiers,
and via MEC-weights was not sufficient to influence the final decision of the meta-learner.
And even though application of MEC-weights to the meta-inputs makes the meta-level cost-sensitive,
the performance of this method is inferior to unweighted stacking models.
We can hypothesise that it may be different if the meta-learner used misclassification costs internally, but
this questions is left for future research.

\paragraph*{Limitations}

Our current work is not without limitations, which we address below.
The choice of the algorithms to be used in stacking is likely to impact its performance.
In order to keep our experiments manageable, we limited ourselves to algorithms used previously 
in the cost-sensitive stacking literature. No parameter tuning was performed to preserve the same base-classifier 
composition across domains.
Those familiar with SVM classifiers could remark that not tuning this algorithm is a mistake.
We are aware of this limitation, which resulted in possibly poor comparative performance of SVM-based ensembles, however
the purpose of the work was to ensure that the ensembles compared differ in only one thing, which is the inclusion of 
cost-sensitive decision-making into different levels of the ensemble.
We are interested in relative performance of stacking setups, not in optimal performance of every learning algorithm
on every domain.
In order to perform statistical tests, we had to ensure that the classifiers were the same in every ensemble for every dataset,
while parameter tuning will have resulted in different parameter settings on different datasets, which would have 
prevented us from performing statistical comparison.
In future work we may experiment with homogeneous stacking, where the diversity of the ensemble
will be created by hyperparameter tuning of the base classifiers.

\section{Conclusions}\label{sec:Conclusion}

Stacking is a well established state-of-the-art ensemble method, that has been widely applied 
to many application domains.
In this work we provide insights into ways to make stacking cost-sensitive.
We compare 90 stacking models built with 15 different compositions of the stacking ensemble using
6 well known classification algorithms. We evaluate on 12 real-world cost-sensitive problems
with clearly defined, non-synthetic, instance-dependent misclassification costs.
In contrast to the absolute majority of cost-sensitive literature, our experimental results demonstarate 
that for the best results, not one, but two layers of cost-sensitive decision-making are required.

We also found that applying MEC-weights to the training inputs of the level-1 classifier in stacking did
not significantly change the performance of stacking models where the level-1 algorithm applied the default decision threshold
to classify.
Moreover, MEC-weighted stacking models where both levels were cost-sensitive performed worse than the unweighted
stacking of the same type, indicating that two levels of cost-sensitivity is sufficient for good performance.

Another contribution of our work is the consolidation of all publically available datasets with record-dependent costs in
one place. In addition to that we derive instance-dependent costs for the well known \emph{credit-g} dataset 
from the UCI repository.
\bibliographystyle{plain}
\bibliography{./paper3}

\appendix
\section{Additional results}\label{apx}
\subsection{Pairwise comparison of stacking setups of type-3 and type-1.}\label{apx:Wilcoxon_type3_vs_type1}
\begin{table}[hbt!]
    \caption{Wilcoxon test statistics (p-values).}\label{tab:wilcoxon_3vs_1}
    \centering
    \begin{adjustbox}{center}
    \begin{tabular}{cllllll}
\toprule
Level-1 algorithm       &  &     type-3 & type-3\_acc &  type-3\_exp &  type-3\_ln &   type-3\_sq \\
\midrule
Ada & type-1 &  0.0 (0.0) &  0.0 (0.0) &   0.0 (0.0) &  0.0 (0.0) &   0.0 (0.0) \\
    & type-1\_acc &  0.0 (0.0) &  0.0 (0.0) &   0.0 (0.0) &  0.0 (0.0) &   0.0 (0.0) \\
    & type-1\_exp &  0.0 (0.0) &  0.0 (0.0) &   0.0 (0.0) &  0.0 (0.0) &   0.0 (0.0) \\
    & type-1\_ln &  0.0 (0.0) &  0.0 (0.0) &   0.0 (0.0) &  0.0 (0.0) &   0.0 (0.0) \\
    & type-1\_sq &  0.0 (0.0) &  0.0 (0.0) &   0.0 (0.0) &  0.0 (0.0) &   0.0 (0.0) \\
\midrule
DT & type-1 &  0.0 (0.0) &  0.0 (0.0) &   0.0 (0.0) &  0.0 (0.0) &   0.0 (0.0) \\
    & type-1\_acc &  0.0 (0.0) &  0.0 (0.0) &   0.0 (0.0) &  0.0 (0.0) &   0.0 (0.0) \\
    & type-1\_exp &  0.0 (0.0) &  0.0 (0.0) &   0.0 (0.0) &  0.0 (0.0) &   0.0 (0.0) \\
    & type-1\_ln &  0.0 (0.0) &  0.0 (0.0) &   0.0 (0.0) &  0.0 (0.0) &   0.0 (0.0) \\
    & type-1\_sq &  0.0 (0.0) &  0.0 (0.0) &   0.0 (0.0) &  0.0 (0.0) &   0.0 (0.0) \\
\midrule
KNN & type-1 &  0.0 (0.0) &  0.0 (0.0) &   0.0 (0.0) &  0.0 (0.0) &   0.0 (0.0) \\
    & type-1\_acc &  0.0 (0.0) &  0.0 (0.0) &   0.0 (0.0) &  0.0 (0.0) &   0.0 (0.0) \\
    & type-1\_exp &  0.0 (0.0) &  0.0 (0.0) &   0.0 (0.0) &  0.0 (0.0) &   0.0 (0.0) \\
    & type-1\_ln &  0.0 (0.0) &  0.0 (0.0) &   0.0 (0.0) &  0.0 (0.0) &   0.0 (0.0) \\
    & type-1\_sq &  0.0 (0.0) &  0.0 (0.0) &   0.0 (0.0) &  0.0 (0.0) &   0.0 (0.0) \\
\midrule
LR & type-1 &  0.0 (0.0) &  0.0 (0.0) &   0.0 (0.0) &  0.0 (0.0) &   0.0 (0.0) \\
    & type-1\_acc &  0.0 (0.0) &  0.0 (0.0) &   0.0 (0.0) &  0.0 (0.0) &   0.0 (0.0) \\
    & type-1\_exp &  0.0 (0.0) &  0.0 (0.0) &   0.0 (0.0) &  0.0 (0.0) &   0.0 (0.0) \\
    & type-1\_ln &  0.0 (0.0) &  0.0 (0.0) &   0.0 (0.0) &  0.0 (0.0) &   0.0 (0.0) \\
    & type-1\_sq &  0.0 (0.0) &  0.0 (0.0) &   0.0 (0.0) &  0.0 (0.0) &   0.0 (0.0) \\
\midrule
RF & type-1 &  0.0 (0.0) &  0.0 (0.0) &   0.0 (0.0) &  0.0 (0.0) &   0.0 (0.0) \\
    & type-1\_acc &  0.0 (0.0) &  0.0 (0.0) &   0.0 (0.0) &  0.0 (0.0) &   0.0 (0.0) \\
    & type-1\_exp &  0.0 (0.0) &  0.0 (0.0) &   0.0 (0.0) &  0.0 (0.0) &   0.0 (0.0) \\
    & type-1\_ln &  0.0 (0.0) &  0.0 (0.0) &   0.0 (0.0) &  0.0 (0.0) &   0.0 (0.0) \\
    & type-1\_sq &  0.0 (0.0) &  0.0 (0.0) &   0.0 (0.0) &  0.0 (0.0) &   0.0 (0.0) \\
\midrule
SVM & type-1 &  0.0 (0.0) &  0.0 (0.0) &  8.0 (0.05) &  1.0 (0.0) &  6.0 (0.03) \\
    & type-1\_acc &  0.0 (0.0) &  0.0 (0.0) &  8.0 (0.05) &  1.0 (0.0) &  6.0 (0.03) \\
    & type-1\_exp &  0.0 (0.0) &  0.0 (0.0) &  8.0 (0.05) &  1.0 (0.0) &  6.0 (0.03) \\
    & type-1\_ln &  0.0 (0.0) &  0.0 (0.0) &  8.0 (0.05) &  1.0 (0.0) &  6.0 (0.03) \\
    & type-1\_sq &  0.0 (0.0) &  0.0 (0.0) &  8.0 (0.05) &  1.0 (0.0) &  6.0 (0.03) \\
\bottomrule
\end{tabular}

    \end{adjustbox}
\end{table}   

\newpage
\subsection{Pairwise comparison of stacking setups of type-3 and type-2.}\label{apx:Wilcoxon_type3_vs_type2}
\begin{table}[hbt!]
    \caption{Wilcoxon test statistics (p-values).}
    \centering
    \begin{adjustbox}{center}
    \begin{tabular}{cllllll}
\toprule
Level-1 algorithm       &  &     type-3 & type-3\_acc &  type-3\_exp &  type-3\_ln &   type-3\_sq \\
\midrule
Ada & type-2 &   0.0 (0.0) &    0.0 (0.0) &   3.0 (0.01) &   3.0 (0.01) &    0.0 (0.0) \\
    & type-2\_acc &   0.0 (0.0) &    0.0 (0.0) &   3.0 (0.01) &   3.0 (0.01) &    0.0 (0.0) \\
    & type-2\_exp &   0.0 (0.0) &    0.0 (0.0) &    0.0 (0.0) &    0.0 (0.0) &    0.0 (0.0) \\
    & type-2\_ln &   0.0 (0.0) &    0.0 (0.0) &    0.0 (0.0) &    0.0 (0.0) &    0.0 (0.0) \\
    & type-2\_sq &   0.0 (0.0) &    0.0 (0.0) &    0.0 (0.0) &    0.0 (0.0) &    0.0 (0.0) \\
\midrule
DT & type-2 &   0.0 (0.0) &   4.0 (0.01) &   3.0 (0.01) &   3.0 (0.01) &   3.0 (0.01) \\
    & type-2\_acc &   0.0 (0.0) &   3.0 (0.01) &   3.0 (0.01) &   3.0 (0.01) &   3.0 (0.01) \\
    & type-2\_exp &   0.0 (0.0) &   3.0 (0.01) &   2.0 (0.01) &    0.0 (0.0) &   3.0 (0.01) \\
    & type-2\_ln &   0.0 (0.0) &    0.0 (0.0) &    0.0 (0.0) &    0.0 (0.0) &    0.0 (0.0) \\
    & type-2\_sq &   0.0 (0.0) &   4.0 (0.01) &   3.0 (0.01) &   3.0 (0.01) &   3.0 (0.01) \\
\midrule
KNN & type-2 &  7.0 (0.04) &   7.0 (0.04) &   7.0 (0.04) &   7.0 (0.04) &   8.0 (0.05) \\
    & type-2\_acc &  7.0 (0.04) &   7.0 (0.04) &   7.0 (0.04) &   7.0 (0.04) &   7.0 (0.04) \\
    & type-2\_exp &  7.0 (0.04) &   7.0 (0.04) &   7.0 (0.04) &   7.0 (0.04) &   8.0 (0.05) \\
    & type-2\_ln &  5.0 (0.02) &   7.0 (0.04) &   7.0 (0.04) &   7.0 (0.04) &   7.0 (0.04) \\
    & type-2\_sq &  7.0 (0.04) &   7.0 (0.04) &   7.0 (0.04) &   7.0 (0.04) &   8.0 (0.05) \\
\midrule
LR & type-2 &   0.0 (0.0) &    0.0 (0.0) &    0.0 (0.0) &    0.0 (0.0) &    0.0 (0.0) \\
    & type-2\_acc &   0.0 (0.0) &    0.0 (0.0) &    0.0 (0.0) &    0.0 (0.0) &    0.0 (0.0) \\
    & type-2\_exp &   0.0 (0.0) &    0.0 (0.0) &    0.0 (0.0) &    0.0 (0.0) &    0.0 (0.0) \\
    & type-2\_ln &   0.0 (0.0) &    0.0 (0.0) &    0.0 (0.0) &    0.0 (0.0) &    0.0 (0.0) \\
    & type-2\_sq &   0.0 (0.0) &    0.0 (0.0) &    0.0 (0.0) &    0.0 (0.0) &    0.0 (0.0) \\
\midrule
RF & type-2 &   0.0 (0.0) &   3.0 (0.01) &   4.0 (0.01) &   3.0 (0.01) &   4.0 (0.01) \\
    & type-2\_acc &   0.0 (0.0) &   3.0 (0.01) &   3.0 (0.01) &   3.0 (0.01) &   3.0 (0.01) \\
    & type-2\_exp &   0.0 (0.0) &   4.0 (0.01) &   4.0 (0.01) &   3.0 (0.01) &   4.0 (0.01) \\
    & type-2\_ln &   0.0 (0.0) &   4.0 (0.01) &   4.0 (0.01) &   3.0 (0.01) &   4.0 (0.01) \\
    & type-2\_sq &   0.0 (0.0) &   3.0 (0.01) &   3.0 (0.01) &   3.0 (0.01) &   4.0 (0.01) \\
\midrule
SVM & type-2 &  6.0 (0.03) &  13.0 (0.16) &  25.0 (0.85) &  13.0 (0.16) &  20.0 (0.49) \\
    & type-2\_acc &  6.0 (0.03) &  14.0 (0.19) &  24.0 (0.77) &  14.0 (0.19) &  20.0 (0.49) \\
    & type-2\_exp &  6.0 (0.03) &  13.0 (0.16) &  24.0 (0.77) &  13.0 (0.16) &  19.0 (0.43) \\
    & type-2\_ln &  6.0 (0.03) &  13.0 (0.16) &  25.0 (0.85) &  13.0 (0.16) &  20.0 (0.49) \\
    & type-2\_sq &  6.0 (0.03) &  13.0 (0.16) &  25.0 (0.85) &  13.0 (0.16) &  19.0 (0.43) \\
\bottomrule
\end{tabular}

    \end{adjustbox}
\end{table}   



\end{document}